\documentclass[11pt]{article}

\usepackage[final]{acl}

\usepackage{times}
\usepackage{latexsym}

\usepackage[T1]{fontenc}

\usepackage[utf8]{inputenc}

\usepackage{microtype}

\usepackage{inconsolata}

\usepackage{graphicx}
\usepackage{amsmath}
\usepackage{amssymb}
\usepackage{algorithm}
\usepackage{algpseudocode}
\usepackage{booktabs}
\usepackage{array}
\usepackage{multirow}

\author{
\textbf{Myunghoon Kang}\textsuperscript{1}\thanks{These authors contributed equally to this work.},
\textbf{Jungseob Lee}\textsuperscript{1}\footnotemark[1],
\textbf{Jaehyung Seo}\textsuperscript{3},
\textbf{Heuiseok Lim}\textsuperscript{1,2}\thanks{Corresponding author} \\
\textsuperscript{1}Department of Computer Science and Engineering, Korea University, \\
\textsuperscript{2}Human-inspired AI Research \\
\textsuperscript{3}Department of Computer Science and Engineering, Konkuk University\\
\textsuperscript{1,2}\texttt{\{chaos8527, omanma1928, limhseok\}@korea.ac.kr}, \textsuperscript{3}\texttt{seojae777@konkuk.ac.kr}
}

\newcommand{\thinkfuse}{\textsc{ThinkFuse}}

\newcommand{\deltagain}[1]{\textcolor{green!50!black}{\scriptsize\,(#1)}}
\newcommand{\deltaloss}[1]{\textcolor{red}{\scriptsize\,(#1)}}
\newcommand{\deltaneutral}[1]{\textcolor{gray!70!black}{\scriptsize\,(#1)}}
\newcommand{\truncmark}{\textsuperscript{\textdagger}}
\newcommand{\segpreview}{T}

\title{\raisebox{-0.62em}{\includegraphics[height=2.4em]{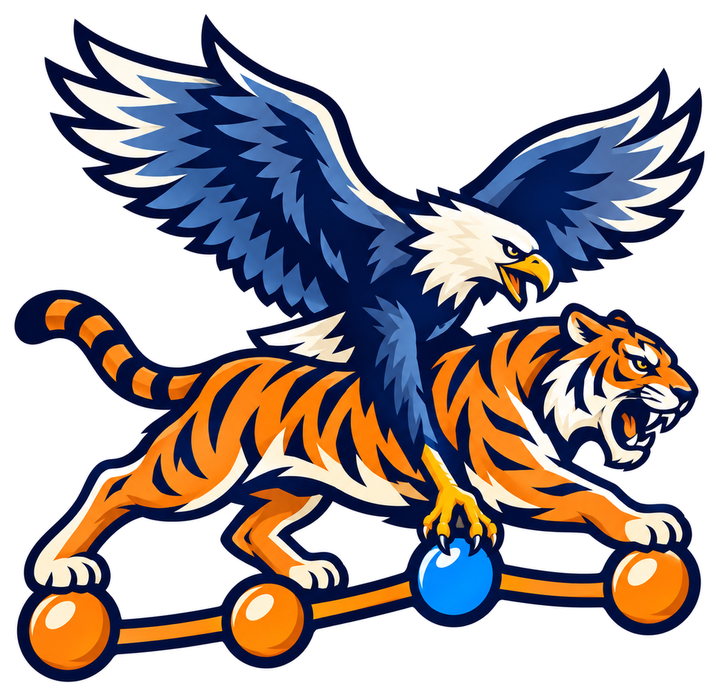}}\hspace{0.28em}ThinkFuse: Trajectory-Aware Test-Time Fusion for \\Small Reasoning Models}

\begin{document}
\maketitle
\begin{abstract}
Small reasoning models (SRMs) have shown strong performance on complex reasoning tasks by generating extended chain-of-thought trajectories, but they often fail to recover once their reasoning enters an erroneous path. Existing test-time fusion methods rely on local fusion signals to determine when to trigger fusion, which can be misled by transient uncertainty fluctuations and may reinforce unstable reasoning trajectories. We propose \thinkfuse{}, a training-free test-time fusion framework that selectively intervenes in unreliable reasoning segments. \thinkfuse{} compares segment-level uncertainty shifts with trajectory-level uncertainty trends to identify unstable reasoning points and fuse auxiliary reasoning paths into the primary model's trajectory. Extensive experiments demonstrate that \thinkfuse{} outperforms baselines on mathematical and knowledge-intensive reasoning benchmarks, with consistent gains across model-family combinations, and remains robust with a smaller primary model. Our analysis shows that \thinkfuse{} requires fewer fusion triggers and generates fewer tokens, highlighting the efficiency of selective triggering. Our code is available at \url{https://github.com/js-lee-AI/ThinkFuse}.
\end{abstract}

\section{Introduction}

\begin{figure*}[t]
\centering
\includegraphics[width=\textwidth]{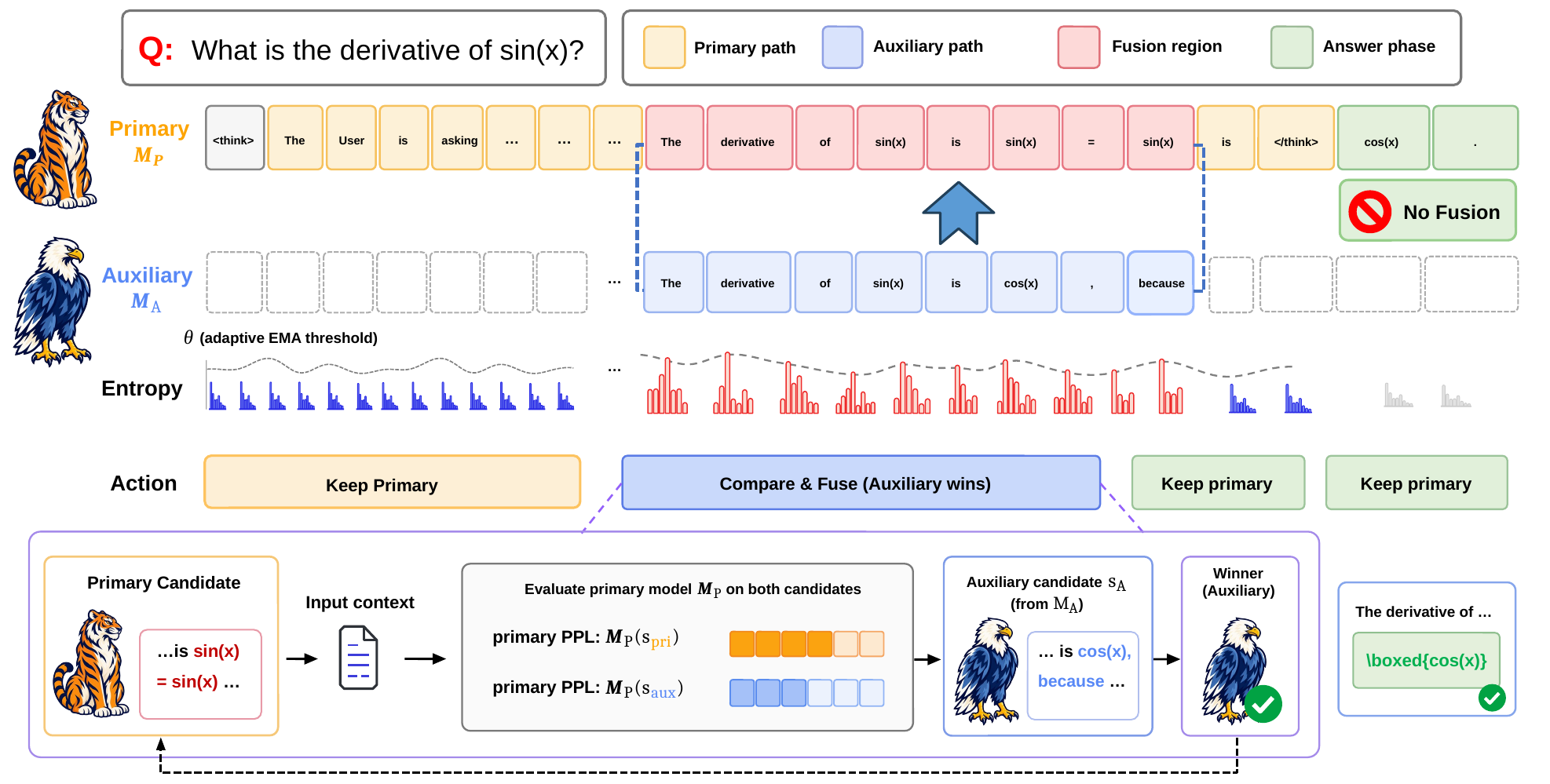}
\caption{Overview of \thinkfuse{} framework. \thinkfuse{} monitors the segment-level uncertainty of the primary model $M_p$ using an adaptive threshold $\theta$. Upon detecting uncertainty spikes, it invokes an auxiliary model $M_a$ to generate alternative reasoning steps. Both trajectory candidates are then evaluated for compatibility by $M_p$ via perplexity ($\text{PPL}$), fusing the most compatible segment into the context. This procedure iterates until the reasoning reaches a terminal state.}
\label{fig1: main_figure}
\end{figure*}

Large Reasoning Models (LRMs) have demonstrated strong performance in complex problem-solving tasks due to their extensive chain-of-thought reasoning~\cite{yang2025qwen3, deepseek2025r1, jiang2025large, wang2026survey}. By generating exploratory and self-reflective reasoning traces within thinking blocks (e.g., \texttt{<think>...</think>}), these models make intermediate reasoning explicit, thereby improving their ability to solve tasks that require multi-step inference, planning, and verification~\cite{shao2024deepseekmath, ding2025multi, yu2025dapo}. While LRMs have established strong baselines across diverse reasoning-intensive domains~\cite{XU2025101370}, small reasoning models (SRMs) struggle to generalize effectively to mathematical and coding tasks~\cite{zhuang2025effective, Souza2025CodeGW}. Scaling down reasoning models while preserving their reasoning capabilities thus remains a critical challenge~\cite{tian2025beyond, chen-etal-2025-unveiling-key, lee2026answer}.

To mitigate this reasoning gap without additional training, test-time fusion has emerged as a promising direction that combines complementary reasoning traces from multiple models~\cite{halim2025study, guo2025stylebench, liu2025coolfusion, yao2025determinethenensemble, xu-etal-2025-hit, adafuse2026}. Previous works have primarily focused on fusion granularity, ranging from post-generation sample-level aggregation~\cite{wang2023selfconsistency,yao2025determinethenensemble} to token or span-level fusion during generation~\cite{liu2025coolfusion, xu-etal-2025-hit}. However, the triggering policy of fusion remains underexplored: when should fusion be initiated, and what evidence should justify the intervention? Existing methods rely on local uncertainty signals at specific decoding steps or predefined fusion points, which fail to capture how uncertainty evolves across the reasoning trajectory. As a result, they may confuse transient uncertainty with reasoning instability, causing premature, delayed, or unnecessary fusion~\cite{li2026offtrajectory, sun2026llmreasoningtrajectoriesstepspecific}.

In this paper, we present \thinkfuse{}, a novel reasoning-path fusion framework for reasoning models. \thinkfuse{} compares segment-level uncertainty shifts with trajectory-level uncertainty trends to identify unstable reasoning points and fuse auxiliary reasoning paths into the primary model's trajectory. Specifically, it invokes the auxiliary model only when the current primary model's reasoning segment exhibits elevated uncertainty relative to the model's recent uncertainty profile. To this end, we propose the Exponentially Weighted Causal Aggregation (EWCA) metric to aggregate token-level uncertainty, which captures intra-segment uncertainty dynamics. ThinkFuse then compares the resulting segment uncertainty with an adaptive threshold estimated from the primary model's historical uncertainty statistics and cumulative fusion frequency. This design regularizes fusion behavior by reducing unnecessary intervention while reserving auxiliary reasoning for uncertain segments. In summary, our paper makes the following contributions:
\begin{itemize}
    \item We propose \thinkfuse{}, a reasoning-model-oriented test-time fusion framework that identifies unreliable reasoning segments using an EWCA-based uncertainty metric and adaptive threshold.

    \item Through extensive experiments on mathematical and knowledge-intensive reasoning benchmarks, we observe that \thinkfuse{} outperforms test-time scaling and test-time fusion baselines, with consistent gains across model-family combinations, and remains robust with a smaller primary model.

    \item We analyze the behavior of \thinkfuse{} and show that it achieves these gains with fewer fusion actions and fewer generated tokens, demonstrating the effectiveness and efficiency of selective fusion triggering for reasoning models.
\end{itemize}

\section{Related Work}
\label{sec:related}
\begin{table*}[t]
\centering
\small
\setlength{\tabcolsep}{5pt}
\renewcommand{\arraystretch}{1.08}
\resizebox{\textwidth}{!}{%
\begin{tabular}{@{}llcccl@{}}
\toprule
\textbf{Method} & \textbf{Intervention unit} & \textbf{Selective trigger} & \textbf{Multi-model fusion} & \textbf{Trajectory-aware} & \textbf{Decision basis} \\
\midrule
Self-consistency~\citep{wang2023selfconsistency} & answer & $\times$ & $\times$ & $\times$ & post-hoc majority voting \\
Cool-Fusion~\citep{liu2025coolfusion} & segment & $\times$ & $\checkmark$ & $\times$ & fuses at every segment \\
AdaFuse~\citep{adafuse2026} & word / span & $\checkmark$ & $\checkmark$ & $\times$ & local token confidence \\
MUR~\citep{yan-etal-2026-mur} & step & $\checkmark$ & $\times$ & $\checkmark$ & single-model momentum uncertainty \\
SpecReason~\citep{specreason} & step & $\times$ & $\checkmark$ & $\times$ & verifies every drafted step \\
\midrule
\thinkfuse{} (ours) & segment & $\checkmark$ & $\checkmark$ & $\checkmark$ & trajectory-calibrated segment uncertainty \\
\bottomrule
\end{tabular}}
\caption{Triggering policies of test-time methods. Intervention unit denotes the granularity at which each method acts on the reasoning process, Selective trigger denotes whether auxiliary computation is invoked only at selected points, Multi-model fusion denotes whether a second model contributes to the trajectory, and Trajectory-aware denotes whether the decision is calibrated against the history of the ongoing trajectory.}
\label{tab:method_comparison}
\end{table*}

\subsection{Test-Time Ensembling and Fusion for Large Language Models}
Test-time ensembling offers a training-free paradigm to enhance the reasoning of Large Language Models (LLMs) by aggregating diverse reasoning candidates from complementary model behaviors. Previous works predominantly optimize fusion \emph{granularity}, spanning macro-level interventions, such as answer-level consensus~\cite{wei2022chain,wang2023selfconsistency}, response-level ranking~\cite{jiang-etal-2023-llm}, and token-level aggregation~\cite{yao2025determinethenensemble}, to intermediate structures like segments~\cite{liu2025coolfusion}, spans~\cite{xu-etal-2025-hit}, and adaptive units~\cite{adafuse2026}. 

However, these granularity-centric approaches overlook the temporal dynamics of ongoing reasoning trajectories, specifically \emph{when} to invoke auxiliary computation. \textsc{ThinkFuse} addresses this gap through a trajectory-aware fusion framework, dynamically triggering auxiliary reasoning by calibrating local segment-level uncertainty against the primary model's historical uncertainty statistics.

\subsection{Large Model Guided Reasoning Correction and Collaboration}
Previous works leverage LLMs to enhance the reasoning capabilities of smaller models, either through offline chain-of-thought supervision~\cite{ranaldi-freitas-2024-aligning} or as external verifiers during online self-correction~\cite{zhang-etal-2024-small}. Beyond training-time guidance, recent works explore test-time collaboration via model routing. For instance, \citet{gupta2024language} employ uncertainty-based rules to defer generations to stronger models, and CITER~\cite{zheng2025citer} utilizes a trainable token-level router to balance generation quality and inference costs. However, these paradigms are primarily designed for passive cost-aware routing mechanisms, rather than active interventions. In contrast, \textsc{ThinkFuse} prioritizes on-the-fly reasoning-path fusion, dynamically invoking auxiliary models only when local segment-level uncertainty indicates a deviation in the primary model's ongoing trajectory.

\subsection{Uncertainty-Guided Test-Time Scaling}
Recent works leverage model uncertainty to modulate test-time scaling. Certaindex~\cite{certaindex} estimates the certainty of reasoning progress to schedule the inference budget, and DeepConf~\cite{deepconf} filters and weights sampled traces by confidence within parallel scaling. MUR~\cite{yan-etal-2026-mur} tracks momentum-smoothed step-level uncertainty within a single trajectory and allocates extra computation to uncertain steps. SpecReason~\cite{specreason} instead lets a small model draft each step and a larger model verify every draft through prompting-based scoring. Rather than allocating computation within a single model or verifying every step, \thinkfuse{} employs uncertainty to trigger selective fusion. By calibrating segment-level uncertainty against the primary model's trajectory history, it delegates only unreliable segments to an auxiliary model and integrates the continuation based on trajectory compatibility.
\section{Methodology}
\label{sec:method}

\subsection{Overview}

We present \textbf{ThinkFuse}, a novel fusion framework for reasoning models. The overall mechanism of ThinkFuse is shown in Figure~\ref{fig1: main_figure} and Algorithm~\ref{alg:thinkfuse}. Given a complex reasoning task $x$, ThinkFuse prompts the primary reasoning model $M_p$ to generate a reasoning path. Leveraging the Exponentially Weighted Causal Aggregation (EWCA) metric and an adaptive threshold $\theta$, the framework monitors uncertainty at the segment level, capturing both internal transition variance and the overall progress of the cumulative trajectory. Once the segment-level uncertainty exceeds the adaptive threshold, ThinkFuse evaluates the primary and auxiliary segments guided by primary-model perplexity under the current context, and extends the reasoning trajectory with the more compatible segment.
Table~\ref{tab:method_comparison} positions \thinkfuse{} against existing test-time scaling methods along three axes. Existing methods either fuse at every segment or rely on local token confidence, and uncertainty-guided scaling methods operate within a single model. \thinkfuse{} is the only method that combines selective triggering, trajectory-aware calibration, and multi-model fusion.

\subsection{Segment-level Fusion Monitoring}
\label{sec:fusion-monitoring}
\begin{algorithm}[t]
\caption{\thinkfuse{}: soft-budget adaptive segment fusion}
\label{alg:thinkfuse}
\footnotesize
\begin{algorithmic}[1]
\Require task $x$, models $M_p, M_a$, hyperparameters $T,\alpha,\beta,\eta,\mu_0,\sigma_0,\tau,\lambda$
\Ensure response $y$
\State $c_p,c_a\gets\textsc{Encode}(x)$; $y\gets\epsilon$; $n,f,R\gets0$
\State $(\mu,\sigma)\gets(\mu_0,\sigma_0)$
\While{not done}
  \State $s_p,\ell_p\gets M_p(c_p,T)$; $s^*\gets s_p$
  \If{$\texttt{</think>}\notin y$}
  \State $U\gets\text{EWCA}(\text{Entropy}(\ell_p))$
  \State $\theta\gets(\mu+\tau\sigma)(1+\lambda R^2)$; update $(\mu,\sigma)$
  \If{$U\ge\theta$}
  \State $s_a\gets M_a(c_a,T)$; $\mathcal{S}\gets\textsc{Align}(s_p,s_a)$
  \State $s^*\gets\arg\min_{s\in\mathcal{S}}\mathrm{PPL}_{M_p}(s\mid c_p)$; $f\gets f+1$
  \EndIf
  \State $n\gets n+1$; $R\gets f/n$
  \EndIf
  \State $y\gets y\oplus s^*$; $(c_p,c_a)\gets\textsc{Sync}(s^*)$
\EndWhile
\State \Return $y$
\end{algorithmic}
\end{algorithm}

Recent studies suggest that reasoning trajectories are organized into thought-level segments, where transition markers such as `wait' or `alternatively' often indicate shifts in reasoning state~\cite{lu2025retro, wang2025thoughts}. Motivated by this view, \thinkfuse{} monitors uncertainty at the segment level rather than relying directly on isolated token-level signals, which can be sensitive to local lexical or discourse fluctuations. Instead of relying on such transition markers as explicit segmentation cues, \thinkfuse{} monitors uncertainty fluctuations of the reasoning trajectory over fixed-length segments. Let $c_k$ denote the reasoning context before generating the $k$-th segment, and $s_k^p = (y_{k,1}^p,\ldots,y_{k,T}^p)$ be the segment generated by the primary model $M_p$ conditioned on $c_k$, where $T$ is the segment length. For each token $y_{k,i}^p$, \thinkfuse{} employs normalized Shannon entropy over the top-$r$ log-probability distribution returned by the primary model:
\begin{equation}
u_{k,i}^p =
-\frac{\sum_{v\in\mathcal{V}_{k,i}}\tilde p_{k,i}(v)\log \tilde p_{k,i}(v)}
{\log |\mathcal{V}_{k,i}|}
\end{equation}
where $\mathcal{V}_{k,i}$ denotes the set of $r$ highest-probability candidate tokens at position $i$, and $\tilde p_{k,i}$ denotes the corresponding distribution renormalized over $\mathcal{V}_{k,i}$.

\paragraph{Intra segment uncertainty calculation}
To aggregate token-level uncertainty into a segment-level signal, ThinkFuse applies EWCA by assigning each token uncertainty $u_{k,i}^p$ a weight based on its relative position and the accumulated uncertainty of preceding tokens:
\begin{equation}
    w_{k,i} = \exp(\alpha \cdot i) \cdot \prod_{j<i}(1 + \beta \cdot u_{k,j}^{p})
\end{equation}
where $\alpha$ controls later-token emphasis and $\beta$ controls uncertainty accumulation. The segment-level uncertainty $U_{k}^{p} $ for the $k$-th segment is computed as
\begin{equation}
    U_k^p =
\frac{\sum_{i=1}^{T_k} w_{k,i} u_{k,i}^p}
{\sum_{i=1}^{T_k} w_{k,i}}.
\end{equation}

\paragraph{Inter-Segment Adaptive Thresholding}
The same uncertainty value can have varying implications depending on the model's recent reasoning trajectory. To this end, \thinkfuse{} employs a triggering policy with an adaptive threshold $\theta_k$. At each step $k$, given the primary model's segment-level uncertainty $U_k^p$, the historical trajectory's mean uncertainty $\mu_k$ and deviation scale $\sigma_k$ are iteratively updated via an exponential moving average:
\begin{equation}
    \mu_k = (1-\eta)\mu_{k-1} + \eta U_{k}^{p}
\end{equation}
\begin{equation}
    \sigma_k = (1-\eta)\sigma_{k-1} + \eta |U_{k}^{p} - \mu_{k-1}|
\end{equation}
\begin{equation}
    \theta_k = (\mu_{k-1}+ \tau \cdot \sigma_{k-1}) \cdot (1 + \lambda \cdot R_{k-1}^2)
\end{equation}

where $\eta \in (0,1)$ is the decay factor and $\tau$ controls the deviation margin. To budget the overall fusion, we introduce $R_{k-1}$ as the cumulative fusion ratio scaled by soft budget parameter $\lambda$. This quadratic modulation penalizes excessive fusion by increasing the threshold as fusions accumulate. By triggering intervention only when $U_k^p \ge \theta_k$, \thinkfuse{} effectively isolates and targets localized uncertainty spikes.

\subsection{Trajectory compatibility scoring}
\label{sec:scoring}
When the fusion trigger activates, ThinkFuse invokes the auxiliary reasoning model $M_a$ to generate an alternative segment $s_k^a$ conditioned on the current context $c_k$. It then treats $s_k^p$ and $s_k^a$ as candidate continuations and uses the primary model $M_p$ as a trajectory-compatibility scorer:
\begin{equation}
\ell_k(s) = \mathrm{PPL}_{M_p}(s \mid c_k),
\end{equation}
for each candidate segment, the selected segment is then defined as
\begin{equation}
s_k^\star =
\begin{cases}
s_k^p, & U_k^p < \theta_k\\
\arg\min\limits_{s \in \{s_k^p, s_k^a\}} \ell_k(s), & U_k^p \ge \theta_k,
\end{cases}
\end{equation}
Then, ThinkFuse appends the selected segment to update the reasoning context:
\begin{equation}
c_{k+1} = c_k \oplus s_k^\star
\end{equation}
The primary model then continues generation from the updated context:
\begin{equation}
s_{k+1}^p \sim M_p(\cdot \mid c_{k+1})
\end{equation}
and this procedure repeats until the reasoning trajectory reaches a terminal
state.

\section{Experimental Setup}
\label{sec:setup}

\subsection{Models}
We conduct experiments using six open-weight reasoning language models: Qwen3-4B~\cite{yang2025qwen3}, Qwen3-1.7B~\cite{yang2025qwen3}, DeepSeek-R1-Distill-Qwen-1.5B~\cite{deepseek2025r1}, Ministral-3B-R~\cite{liu2026ministral}, EXAONE-Deep-2.4B~\cite{bae2025exaone}, and EXAONE-4.0-1.2B~\cite{bae2025exaone4}. These models span multiple model families and cover a practical small-scale regime from 1.2B to 4B parameters, allowing us to evaluate whether \textsc{ThinkFuse} can improve reasoning without relying on larger proprietary models. The model pool includes both reasoning-specialized models, such as Ministral-3B-R and EXAONE-Deep-2.4B, and hybrid or instruction-following models with reasoning-oriented generation capabilities, such as Qwen3-4B, Qwen3-1.7B, and EXAONE-4.0-1.2B. This diversity is important for \textsc{ThinkFuse}, as the framework is designed to exploit complementary reasoning trajectories across models rather than assuming homogeneous decoding behaviors or failure modes.

\subsection{Baselines}
We compare \textsc{ThinkFuse} against three types of baselines. 
\paragraph{Standalone} evaluates each reasoning model independently, allowing us to assess whether \textsc{ThinkFuse} provides gains beyond each model's original reasoning capability.\ 
\paragraph{Test-time fusion methods} include \textbf{Cool-Fusion}~\citep{liu2025coolfusion}, a segment-level fusion method that selects model-generated continuations using perplexity-based scoring, and \textbf{AdaFuse}~\citep{adafuse2026}, an adaptive ensemble decoding method that dynamically adjusts word/span-level fusion based on model confidence. 

\paragraph{Test-time compute baselines} 
\textbf{Self-consistency} ~\citep{wang2023selfconsistency} generates $K\in\{3,5\}$ sampled reasoning paths with Qwen3-4B. Each final answer is normalized using the same answer parser as the main evaluation, and the prediction is determined by majority voting over normalized answers. If no normalized answer obtains a majority, we use the first generated sample as the final prediction.

\paragraph{\textsc{ThinkFuse}} evaluates our reasoning-path fusion framework under both cross-family and same-family model pairs, including configurations with either a larger or smaller primary model. 

\begin{table*}[t!]
\centering
\small
\setlength{\tabcolsep}{4pt}
\resizebox{0.9\textwidth}{!}{%
\begin{tabular}{lccccc}
\toprule
\textbf{Method} & \textbf{MATH-500} & \textbf{GSM8K} & \textbf{AIME24} & \textbf{GPQA} & \textbf{NQ-Open} \\
\midrule
\multicolumn{6}{l}{\emph{Standalone models}} \\
Qwen3-4B  & 89.0 & 92.0 & 53.3 & 54.0 & 29.5 \\
Qwen3-1.7B\textsuperscript{\ddag}  & 83.0 & 90.0 & 33.3 & 39.4 & 28.0 \\
Ministral & 32.5 & 67.0 & 0.0\truncmark  & 25.8 & 24.0 \\
EXAONE-4.0  & 57.0 & 85.0 & 3.3\truncmark & 38.9 & 22.0 \\
\midrule
\multicolumn{6}{l}{\emph{Test-time compute baseline: Qwen3-4B}} \\
Self-consistency ($K=3$) & 89.5\deltagain{+0.5} & 94.5\deltagain{+2.5} & 53.3\deltaneutral{+0.0} & 51.0\deltaloss{$-$3.0} & 39.0\deltagain{+9.5} \\
Self-consistency ($K=5$) & 88.0\deltaloss{$-$1.0} & 94.5\deltagain{+2.5} & 53.3\deltaneutral{+0.0} & 50.0\deltaloss{$-$4.0} & 39.0\deltagain{+9.5} \\
\midrule
\multicolumn{6}{l}{\emph{Test-time fusion baselines: Qwen3-4B $\times$ Ministral}} \\
Cool-Fusion  & 31.0\deltaloss{$-$58.0} & 66.0\deltaloss{$-$26.0} & 0.0\truncmark\deltaloss{$-$53.3} & 11.6\deltaloss{$-$42.4} & 0.5\deltaloss{$-$29.0} \\
AdaFuse  & 45.5\deltaloss{$-$43.5} & 72.6\deltaloss{$-$19.4} & 33.3\deltaloss{$-$20.0} & 28.8\deltaloss{$-$25.2} & 38.5\deltagain{+9.0} \\
\midrule
\multicolumn{6}{l}{\emph{\thinkfuse{}}} \\
Qwen3-4B $\times$ Ministral  & 90.0\deltagain{+1.0} & \textbf{96.9}\deltagain{+4.9} & \textbf{73.3}\deltagain{+20.0} & \textbf{64.6}\deltagain{+10.6} & \textbf{39.5}\deltagain{+10.0} \\
Qwen3-4B $\times$ Qwen3-1.7B  & \textbf{91.0}\deltagain{+2.0} & 94.0\deltagain{+2.0} & 70.0\deltagain{+16.7} & 55.1\deltagain{+1.1} & 37.0\deltagain{+7.5} \\
Qwen3-4B $\times$ Qwen3-4B  & 75.0\deltaloss{$-$14.0} & 90.0\deltaloss{$-$2.0} & 40.0\deltaloss{$-$13.3} & 28.3\deltaloss{$-$25.7} & 38.0\deltagain{+8.5} \\
Qwen3-1.7B $\times$ EXAONE-4.0 & 85.0\deltagain{+2.0} & 90.5\deltagain{+0.5} & 30.0\deltaloss{$-$3.3} & 53.5\deltagain{+14.1} & 28.0\deltaneutral{+0.0} \\
\bottomrule
\end{tabular}
}
\caption{Main benchmark results. Accuracy (\%) for standalone models, test-time baselines, and \thinkfuse{} across five benchmarks under a 16K-token budget. Values in parentheses indicate the signed accuracy difference, in percentage points, relative to the standalone performance of the primary model, with green indicating gains and red indicating degradation. The $\dagger$ symbol marks AIME24 results affected by length-limit or final-answer truncation. The $\ddag$ symbol marks the standalone Qwen3-1.7B evaluated with the Qwen3 recommended top-$k$ of 20, whereas all other rows use the shared decoding setting. Additional experimental details, including the decoding settings and remaining cross-family and same-family configurations, are presented in Appendix~\ref{app:imp_details} and~\ref{app:full_results}.}
\label{tab:main_results}
\end{table*}

\subsection{Datasets}
We evaluate \textsc{ThinkFuse} on two categories of reasoning benchmarks. To assess complex mathematical reasoning, we use MATH-500~\citep{hendrycks2021math}, GSM8K~\citep{cobbe2021gsm8k}, and AIME 2024\footnote{\url{https://huggingface.co/datasets/Maxwell-Jia/AIME_2024}}, which respectively evaluate competition-level mathematical problem solving, multi-step grade-school arithmetic reasoning, and advanced high-school competition mathematics. To assess knowledge-intensive reasoning capability, we use GPQA-Diamond~\citep{rein2024gpqa} and NQ-Open~\citep{kwiatkowski2019nq}. GPQA-Diamond evaluates expert-level scientific reasoning across biology, physics, and chemistry, whereas NQ-Open evaluates open-domain factual question answering from naturally occurring questions. 

\subsection{Implementation Details}
All experiments are implemented with vLLM~\cite{kwon2023vllm} on two NVIDIA A100-80GB GPUs. For efficient test-time fusion, ThinkFuse executes auxiliary candidate generation in parallel with primary-model generation whenever applicable. We use publicly available Hugging Face checkpoints~\cite{wolf-etal-2020-transformers} for all primary and auxiliary models. We report accuracy on all benchmarks, evaluating GPQA and NQ-Open with the LM Evaluation Harness~\cite{eval-harness} and MATH-500, GSM8K, and AIME-24 following Qwen2.5-Math~\cite{yang2024qwen2math}. All methods share a 16,384 token generation budget and one decoding configuration, except that the standalone Qwen3-1.7B uses the Qwen3 recommended top-$k$ of 20, and responses without a parseable final answer within the budget are scored as incorrect. We evaluate each method on a fixed random subsample of each benchmark using a consistent sampling protocol across methods. Detailed experimental setups are summarized in Appendix~\ref{app:imp_details}.

\section{Main Results}
\label{sec:results}
Table~\ref{tab:main_results} reports the accuracy of standalone models, existing test-time compute baselines, test-time fusion baselines, and \textsc{ThinkFuse} across five benchmarks. For brevity throughout the paper, we abbreviate the model names as follows: Ministral (Ministral-3B-R) and EXAONE-4.0 (EXAONE-4.0-1.2B). Among standalone models, Qwen3-4B shows the strongest and most stable performance across benchmarks, providing a strong baseline for test-time fusion. Despite this strong baseline, \textsc{ThinkFuse} with Qwen3-4B as the primary model improves over its standalone performance on every benchmark, whether the auxiliary is a cross-family model or a smaller model from the same family. In particular, Qwen3-4B $\times$ Ministral achieves the best result on four out of five benchmarks, with large gains on AIME24 (+20.0), GPQA (+10.6), and NQ-Open (+10.0). The additional cross-family pairs with EXAONE-Deep-2.4B and DeepSeek-R1-Distill as auxiliaries yield the same pattern (Appendix~\ref{app:full_results}), suggesting that \textsc{ThinkFuse} can benefit from auxiliary reasoning trajectories even when the auxiliary model is not uniformly stronger than the primary model. Bootstrap confidence intervals in Appendix~\ref{app:reliability} confirm this consistency, as all twenty deltas across the four Qwen3-4B pairs are positive and eleven are significant at the 95\% level.

Conversely, existing test-time fusion baselines show the opposite trend. With the same Qwen3-4B $\times$ Ministral pair, Cool-Fusion drops sharply from 89.0 to 31.0 on MATH-500 and from 54.0 to 11.6 on GPQA, while AdaFuse also underperforms the standalone primary on four out of five benchmarks. Test-time compute through self-consistency provides gains on GSM8K and NQ-Open, but does not improve AIME24 and degrades GPQA, indicating that simply increasing test-time sampling is not sufficient for consistent improvement. Moreover, the redundant same-model variant, Qwen3-4B $\times$ Qwen3-4B, degrades performance on MATH-500, GSM8K, AIME24, and GPQA. These results suggest that fusion must be selective and trajectory-aware, since locally determined or redundant fusion can disrupt the primary model's reasoning process.  

Finally, we examine whether \thinkfuse{} remains beneficial with a smaller primary model by pairing Qwen3-1.7B with EXAONE-4.0. The standalone Qwen3-1.7B is evaluated with the Qwen3 recommended top-$k$ of 20, while \textsc{ThinkFuse} retains the shared decoding setting, so this comparison favors the standalone baseline. Even so, \textsc{ThinkFuse} raises GPQA from 39.4 to 53.5, exceeding both standalone models. On the remaining benchmarks, where EXAONE-4.0 is substantially weaker than the primary, \textsc{ThinkFuse} gains +2.0 on MATH-500 and +0.5 on GSM8K, ties on NQ-Open, and differs by a single problem on AIME24. Thus, this distribution of gains matches the design principle of \thinkfuse{}: complementary auxiliary trajectories are exploited where they exist, while the selective trigger and PPL-based compatibility scoring prevent harmful fusion where they do not. Overall, the results suggest that \thinkfuse{} improves reasoning performance by selectively fusing complementary auxiliary trajectories at uncertain points, yielding more stable and effective test-time fusion than existing baselines. 

\begin{figure}[t]
\centering
\includegraphics[width=\columnwidth]{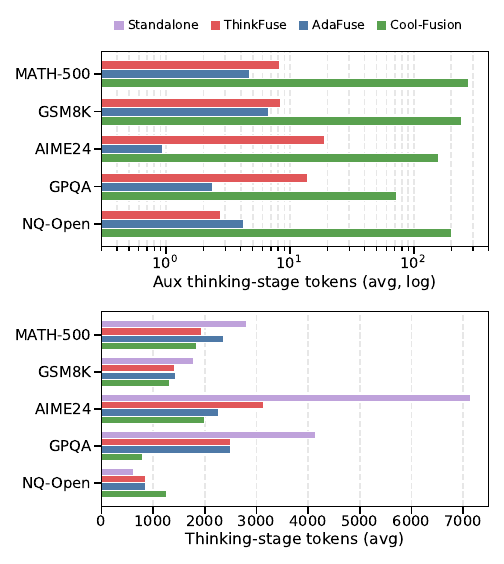}
\caption{Thinking-stage token statistics. Top: Average fused auxiliary tokens in log scale. Bottom: Average total thinking-stage tokens.  All fusion configurations use Qwen3-4B as primary model and Ministral-3B-Reasoning as auxiliary model.}

\label{fig:fusion_rates}
\end{figure}

\section{Analysis}
To understand the sources of \thinkfuse{}'s accuracy gains, we conduct three complementary analyses. First, we examine how fusion behavior relates to downstream performance across methods. Second, we measure the compute efficiency of each test-time fusion method. Third, we test whether \thinkfuse{} remains effective when the auxiliary model is transformed to instruction-tuned variant.

\subsection{Fusion Dynamics and Downstream Performance}
Figure~\ref{fig:fusion_rates} analyzes how different fusion methods use auxiliary reasoning tokens and how this affects the overall thinking-stage length. Despite fusing the most auxiliary tokens, Cool-Fusion yields the shortest thinking-stage outputs.  This suggests that excessive fusion can overwrite or prematurely redirect the primary model's reasoning trajectory, explaining its large performance drops in Table~\ref{tab:main_results}. Meanwhile, AdaFuse uses fewer auxiliary tokens, but still shortens the thinking stage on reasoning-intensive benchmarks such as AIME24 and GPQA, indicating that simply limiting auxiliary usage is insufficient for preserving reasoning quality.

In contrast, \textsc{ThinkFuse} uses auxiliary tokens much more sparsely than Cool-Fusion while maintaining a more substantial reasoning process than the baselines. Combined with the consistent accuracy gains in Table~\ref{tab:main_results}, this analysis suggests that effective test-time fusion requires selective and trajectory-aware intervention, rather than frequent or locally determined auxiliary fusion.

\subsection{Compute Efficiency}
\label{sec:efficiency}
Table~\ref{tab:efficiency} compares the efficiency of the test-time fusion methods on the Qwen3-4B $\times$ Ministral pair, with accuracy averaged over the five benchmarks. \thinkfuse{} achieves the highest accuracy with a TFLOP budget comparable to AdaFuse and lower than that of Cool-Fusion, resulting in the highest accuracy per TFLOP. Meanwhile, \thinkfuse{} exhibits a longer wall-clock time, yet the comparable TFLOPs indicate that this gap arises from the CPU-GPU data transfer during PPL scoring rather than from additional computation. The lower latency of the baselines, in turn, comes at the cost of severe accuracy degradation, leaving \thinkfuse{} as the only fusion method that improves over the standalone primary.

\begin{table}[t]
\centering
\small
\setlength{\tabcolsep}{4pt}
\renewcommand{\arraystretch}{1.08}
\resizebox{\columnwidth}{!}{
\begin{tabular}{@{}lcccc@{}}
\toprule
\textbf{Method} & \textbf{Wall-clock (s)} & \textbf{TFLOPs} & \textbf{Acc.} & \textbf{Acc. / TFLOPs} \\
\midrule
Cool-Fusion & 39.4 & 86 & 21.8 & 0.25 \\
AdaFuse & \textbf{33.9} & \textbf{74} & 43.7 & 0.60 \\
\thinkfuse{} & 68.7 & 77 & \textbf{72.9} & \textbf{0.95} \\
\bottomrule
\end{tabular}
}
\caption{Efficiency comparison of \thinkfuse{} and comparable test-time fusion methods. Wall-clock time and TFLOPs are averaged per example, and accuracy is averaged over the five benchmarks, with the Qwen3-4B $\times$ Ministral pair.}
\label{tab:efficiency}
\end{table}

\subsection{Auxiliary Compatibility Beyond Reasoning Models}
We explore \thinkfuse{}'s extensibility to instruction-tuned and non-thinking auxiliary models. As shown in Table~\ref{tab:it_aux}, these models yield limited, task-dependent gains. On MATH-500, models like Llama-3.2-3B-IT and Qwen3-1.7B (no-think) marginally improve the Qwen3-4B baseline, while others slightly degrade performance. Consequently, while these findings provide partial evidence of extensibility, reasoning-specialized auxiliaries remain fundamentally aligned with \textsc{ThinkFuse}'s core trajectory-fusion mechanism.

Furthermore, on GPQA, all non-reasoning auxiliaries uniformly degrade performance, contrasting sharply with the consistent gains from cross-family reasoning models shown in Table~\ref{tab:main_results} and Appendix~\ref{app:full_results}. This confirms that effective fusion relies not on mere auxiliary diversity, but on the auxiliary's capacity to generate reasoning segments that coherently complement the primary model's ongoing trajectory. Thus, while non-reasoning models demonstrate partial extensibility, reasoning-specialized auxiliaries remain essential to fully leverage \textsc{ThinkFuse}'s trajectory-aware mechanism.

\begin{table}[t]
\centering
\small
\setlength{\tabcolsep}{5pt}
\resizebox{0.95\columnwidth}{!}{%
\begin{tabular}{lcc}
\toprule
\textbf{Auxiliary} & \textbf{MATH-500} & \textbf{GPQA} \\
\midrule
Llama-3.2-3B-IT & 89.5\deltagain{+0.5} & 50.5\deltaloss{$-$3.5} \\
Gemma-2-2B-it & 87.5\deltaloss{$-$1.5} & 51.0\deltaloss{$-$3.0} \\
Mistral-7B-Inst-v0.3 & 84.5\deltaloss{$-$4.5} & 44.9\deltaloss{$-$9.1} \\
Qwen2.5-3B-Inst & 88.0\deltaloss{$-$1.0} & 52.5\deltaloss{$-$1.5} \\
Qwen3-1.7B (no-think) & 89.5\deltagain{+0.5} & 50.0\deltaloss{$-$4.0} \\
\bottomrule
\end{tabular}
}
\caption{Experimental results of pairing Qwen3-4B as a primary model with instruction-tuned or non-thinking auxiliaries. The general performance drop indicates that \thinkfuse{} strictly requires reasoning-specialized models to ensure trajectory compatibility. Values in parentheses show the difference from the standalone baseline.}
\label{tab:it_aux}
\end{table}

\begin{table}[t]
\centering
\small
\setlength{\tabcolsep}{6pt}
\renewcommand{\arraystretch}{1.08}
\resizebox{\columnwidth}{!}{%
\begin{tabular}{lcc}
\toprule
\textbf{Setting} & \textbf{MATH-500} & \textbf{GSM8K} \\
\midrule
\multicolumn{3}{c}{\emph{Fusion budget penalty}} \\
Soft budget (ours, $\lambda{=}50$) & \textbf{90.0}\deltagain{+1.0} & \textbf{96.9}\deltagain{+4.9} \\
No budget penalty ($\lambda{=}0$) & 88.5\deltaloss{$-$0.5} & 90.5\deltaloss{$-$1.5} \\
\midrule
\multicolumn{3}{c}{\emph{Segment length $T$}} \\
$T{=}4$ & 87.0\deltaloss{$-$2.0} & 93.5\deltagain{+1.5} \\
$T{=}8$ (ours) & \textbf{90.0}\deltagain{+1.0} & \textbf{96.9}\deltagain{+4.9} \\
$T{=}16$ & 88.0\deltaloss{$-$1.0} & 93.5\deltagain{+1.5} \\
$T{=}32$ & 88.5\deltaloss{$-$0.5} & 93.5\deltagain{+1.5} \\
\midrule
\multicolumn{3}{c}{\emph{Trajectory compatibility scoring}} \\
Primary PPL (ours) & \textbf{90.0}\deltagain{+1.0} & \textbf{96.9}\deltagain{+4.9} \\
Average PPL & 80.5\deltaloss{$-$8.5} & 92.0\deltaneutral{+0.0} \\
\midrule
\multicolumn{3}{c}{\emph{Token-level uncertainty estimation}} \\
Entropy (ours) & \textbf{90.0}\deltagain{+1.0} & 96.9\deltagain{+4.9} \\
Confidence gap & 87.5\deltaloss{$-$1.5} & \textbf{98.6}\deltagain{+6.6} \\
\bottomrule
\end{tabular}
}
\caption{Impact of individual \thinkfuse{} components on downstream performance. Values in parentheses denote the accuracy delta relative to the standalone baseline. All experiments are conducted with Qwen3-4B as the primary model and Ministral-3B-Reasoning as the auxiliary model.}

\label{tab:ablation}
\end{table}

\begin{table*}[t]
\centering
\small
\setlength{\tabcolsep}{6pt}
\renewcommand{\arraystretch}{1.00}
\resizebox{\textwidth}{!}{%
\begin{tabular}{p{0.47\textwidth}p{0.47\textwidth}}
\toprule
\multicolumn{2}{p{0.94\textwidth}}{\textbf{Shared prefix.} \emph{For AIME-2022 II \#11, both traces begin with the same coordinate setup: $A=(0,0)$, $B=(2,0)$, $D=(p,q)$, $C=(r,s)$, and $M=((2+r)/2,s/2)$ is the midpoint of $BC$. Both traces then try to use the two angle-bisector conditions at $M$.}} \\
\midrule
\textbf{Standalone} & \textbf{\thinkfuse{}} \\
\midrule
\textbf{\textcolor{red!70!black}{Generated trace near the divergence.}}

\vspace{0.60em}
\emph{\(\cdots\) I place the quadrilateral in coordinates and write the distance equations. Since $M$ is the midpoint of $BC$, it should lie on the two angle bisectors. \(\cdots\)}

\vspace{0.70em}
{\ttfamily\textcolor{red!70!black}{the valid solution is b = -7/sqrt(2), a = 7/sqrt(2)}}

{\ttfamily\textcolor{red!70!black}{So, D is at ((7*sqrt(2))/2, -7*sqrt(2)/2).}}

\vspace{0.65em}
\emph{\(\cdots\) I plug this coordinate for $D$ back into the remaining constraints and solve for $C$. \(\cdots\)}

\vspace{0.60em}
{\ttfamily (2+sqrt(2))*y\string^2 + 9*(sqrt(2)+1)*y}

{\ttfamily \hspace{1em} + (22+7*sqrt(2)) = 0}

\vspace{0.60em}
{\scriptsize\ttfamily\textcolor{gray!65!black}{// Comment: this fixes D as if $\angle DAB=45^\circ$. The second angle-bisector condition has not been used, so the trace enters a dead-end quadratic.}}
&
\textbf{\textcolor{green!40!black}{Generated trace near the fusion point.}}

\vspace{0.60em}
\emph{\(\cdots\) I keep $D=(p,q)$ and $C=(r,s)$ symbolic, then express the two angle-bisector conditions as distance-to-line equations. \(\cdots\)}

\vspace{0.60em}
{\ttfamily [Two angle-bisector distance equations from M = (B+C)/2]}

{\ttfamily -8 - 2p + 3r = 0}

\vspace{0.60em}
{\ttfamily\textcolor{green!40!black}{=> r = (8 + 2p)/3}}

{\ttfamily\textcolor{green!40!black}{=> s = 2q/3}}

\vspace{0.60em}
{\ttfamily Substituting these relations keeps both}

{\ttfamily angle-bisector constraints active while the}

{\ttfamily remaining equations are solved.}

\vspace{0.60em}
{\scriptsize\ttfamily\textcolor{gray!65!black}{// Comment: the auxiliary segment keeps $D=(p,q)$ symbolic. It uses both angle-bisector conditions before solving.}}
\\
\addlinespace[0.15em]
\cmidrule(lr){1-2}
{\footnotesize\textbf{Answer:} \textcolor{red!70!black}{$\boxed{0}\ \boldsymbol{\times}$}}
&
{\footnotesize\textbf{Answer:} \textcolor{green!40!black}{$\boxed{180}\ \checkmark$}}
\\
\bottomrule
\end{tabular}
}
\caption{Qualitative comparison on AIME 2022. Standalone denotes Qwen3-4B. \thinkfuse{} uses Qwen3-4B as the primary model and Ministral-3B-Reasoning as the auxiliary model. Both traces start from the same coordinate formulation; the highlighted text marks the standalone step that enters a dead-end quadratic and the auxiliary symbolic step selected by \thinkfuse{}.}
\label{tab:qualitative_case}
\end{table*}

\section{Ablation Study}
Table~\ref{tab:ablation} details the contribution of each algorithmic component within \thinkfuse{} to the overall performance. Throughout these ablations, Qwen3-4B and Ministral-3B-Reasoning are employed as the primary and auxiliary models, respectively, with color-coded values in parentheses denoting the accuracy delta relative to the Qwen3-4B baseline.

\paragraph{Fusion budget penalty ablation}
As shown in Table~\ref{tab:ablation}, the soft budget penalty ($\lambda=50$) outperforms the no-budget variant ($\lambda=0$), which removes the cumulative fusion ratio penalty from the adaptive threshold $\theta_k$. Yielding accuracies of 88.5\% on MATH-500 and 90.5\% on GSM8K, the no-budget variant degrades performance even below the standalone primary model's baseline. These results indicate that the proposed soft budget stabilizes \thinkfuse{} by actively discouraging excessive fusion interventions.

\paragraph{Effects of segment length} 
The segment length $T$ dictates a crucial trade-off in \thinkfuse{}. Our default setting of $T=8$ yields the best results, achieving 90.0\% on MATH-500 and 96.9\% on GSM8K. When the segment is too short ($T=4$), it fails to encapsulate a meaningful reasoning step, causing MATH-500 accuracy to fall below the standalone baseline (87.0\%). Conversely, expanding the segment length ($T=16$ and $T=32$) substantially diminishes the gains on GSM8K and similarly hurts MATH-500. This suggests that excessively long segments delay necessary interventions, allowing incorrect reasoning paths to propagate before fusion occurs.

\paragraph{Effects of compatibility selection rule} 
We investigate PPL scoring rules as a proxy for trajectory compatibility. As shown in Table~\ref{tab:ablation}, relying solely on the Primary PPL yields optimal results (90.0\% on MATH-500, 96.9\% on GSM8K). In contrast, the Average PPL scoring rule drops MATH-500 performance to 80.5\% and limits GSM8K to the standalone Qwen3-4B baseline (92.0\%). These results indicate that blending the auxiliary model's distribution into the primary model's score misaligns the compatibility metric, causing the selected segments to diverge from the primary model's ongoing reasoning path.

\paragraph{Token-level uncertainty estimation}
We compare entropy against a margin-based confidence-gap signal (see Appendix~\ref{app:uncertainty}). Although the confidence gap achieves 98.6\% on GSM8K, it degrades on MATH-500 (87.5\%), falling below the standalone baseline. As entropy consistently preserves performance gains on both datasets, we employ it as our default token-level uncertainty metric.

\section{Case Study}
Table~\ref{tab:qualitative_case} illustrates how \thinkfuse{} corrects intermediate reasoning trajectories using an AIME 2022 example. Despite sharing the initial coordinate setup, the standalone baseline prematurely grounds the coordinates of $D$, failing to satisfy the second angle-bisector condition and trapping the reasoning in a dead-end quadratic, resulting in an incorrect answer. In contrast, \thinkfuse{} triggers an intervention and selects an auxiliary segment that keeps $D$ and $C$ symbolic and successfully enforces both distance-to-line constraints before solving. This case demonstrates that injecting an appropriate symbolic segment preserves the shared mathematical context while effectively redirecting the model from an erroneous path toward the correct solution.

\section{Conclusion}
We presented \textsc{ThinkFuse}, a trajectory-aware test-time fusion framework that selectively intervenes in model reasoning trajectories. By calibrating segment-level uncertainty with historical statistics and applying trajectory compatibility scoring, \textsc{ThinkFuse} integrates auxiliary reasoning only when the primary trajectory is likely to benefit. Across five benchmarks, \textsc{ThinkFuse} consistently outperforms standalone models and existing test-time fusion baselines, showing that its gains come from complementary auxiliary trajectories rather than additional computation alone. These results suggest that effective collaborative reasoning depends on deciding \emph{when} to intervene and \emph{what} to retain, offering a practical, training-free approach for scaling small reasoning models at test time.

\section*{Limitations}
\paragraph{Test-time resource requirements.}
Unlike single-model decoding, \thinkfuse{} requires an execution environment that can serve an auxiliary model together with the primary model. This can increase memory usage and latency. However, \thinkfuse{} intervenes only on selected segments in the explicit thinking stage, and its PPL-based segment selection reuses the current KV cache to avoid an additional full forward pass over the entire context. Quantization, parallel serving, and cache-aware deployment may further reduce practical overhead.

\paragraph{Evaluation scope.}
Our main experiments focus on open-weight reasoning models in the 1.2B--4B range. This setting matches our goal of improving reasoning under limited resources, but additional evaluation is needed to determine whether the same behavior holds for larger models, closed models, and broader model families. We also focus on mathematical, scientific, and knowledge-based reasoning benchmarks; code generation, multilingual reasoning, and long-context tasks remain outside the scope of this study.

\paragraph{Dependence on primary-model quality.}
\thinkfuse{} is designed for settings where the primary model can produce coherent reasoning segments but may enter an incorrect reasoning path at uncertain points. If the primary model generates highly unstable or repetitive segments, uncertainty-based triggering may fail to identify useful intervention points. Such cases may require alternative decoding policies or stronger criteria for selecting the primary model.

\paragraph{Auxiliary-model complementarity.}
The benefit of \thinkfuse{} depends on whether the auxiliary model provides reasoning segments that complement the primary model. If the auxiliary model shares similar failure modes with the primary model, or if its strengths do not match the target task, fusion may not always improve performance. Future work can reduce this dependence through task-specific auxiliary selection or model-pool routing.

\paragraph{Explicit reasoning spans and tokenizer alignment.}
Because \thinkfuse{} intervenes at the segment level inside the reasoning stage, it requires identifying the explicit reasoning span and aligning text segments generated under different tokenizers. Our implementation uses \texttt{<think>} tag normalization and shared-prefix text alignment, but additional engineering may be needed for broader chat templates and tokenizer combinations.

\paragraph{Fixed fusion settings.}
We use a fixed segment length and PPL scoring rule in the main experiments and analyze representative alternatives in ablations. More extensive searches over threshold and budget hyperparameters may yield additional gains. Our focus is instead to test whether uncertainty-guided fusion is effective without heavy task-specific tuning.

\section*{Ethical Statements}
In this section, we discuss the ethical considerations of our work.

\paragraph{(1) Privacy and Languages.} All benchmarks used in our experiments are publicly available academic datasets in English. MATH-500~\citep{hendrycks2021math}, GSM8K~\citep{cobbe2021gsm8k}, and AIME 2024 consist of competition-level mathematical problems, multi-step grade-school arithmetic, and advanced high-school competition mathematics, respectively. GPQA-Diamond~\citep{rein2024gpqa} provides graduate-level scientific reasoning questions in biology, physics, and chemistry written by domain experts, and NQ-Open~\citep{kwiatkowski2019nq} contains open-domain factual questions derived from naturally occurring Google search queries with Wikipedia-grounded answers. Since each benchmark is built from curated academic problems or factual question answering, there is minimal risk of including personally identifiable information or offensive content. 
\paragraph{(2) Licenses and terms.} The reasoning models we evaluate are released under open-weight licenses that permit research use: Qwen3-4B, Qwen3-1.7B~\cite{yang2025qwen3}, and Ministral-3B-Reasoning~\cite{liu2026ministral}  are distributed under the Apache 2.0 license, DeepSeek-R1-Distill~\cite{deepseek2025r1} under the MIT license, and EXAONE-Deep-2.4B~\cite{bae2025exaone} / EXAONE-4.0-1.2B~\cite{bae2025exaone4} under their respective model licenses for research purposes. We primarily employ vLLM~\cite{kwon2023vllm} (Apache 2.0), Hugging Face Transformers~\cite{wolf-etal-2020-transformers} (Apache 2.0), and the LM Evaluation Harness~\cite{eval-harness} (MIT) for model serving and evaluation, all of which are available under permissive open-source licenses. 
\paragraph{(3) Intended use.} Our proposed \thinkfuse{} framework is intended for research on test-time fusion of small reasoning models, with the goal of improving the reliability of their reasoning trajectories without additional training. By selectively invoking auxiliary reasoning only at uncertain segments, \thinkfuse{} can serve as a basis for studying \emph{when} and \emph{what} to fuse during multi-model reasoning collaboration; we will release our code and evaluation scripts to support reproducibility and follow-up research on trajectory-aware fusion for resource-constrained reasoning.

 \paragraph{(4) Use of AI assistants.} During the preparation of this manuscript, we used AI assistants for grammar correction and minor language polishing of text originally drafted by the authors. All AI-assisted edits were subsequently reviewed and verified by the authors, who take full responsibility for the final content of the paper.

\section*{Acknowledgements}
This research was supported by Basic Science Research Program through the National Research Foundation of Korea(NRF) funded by the Ministry of Education(NRF-2021R1A6A1A03045425). This work was supported by Institute for Information \& communications Technology Promotion(IITP) grant funded by the Korea government(MSIT) (RS-2024-00398115, Research on the reliability and coherence of outcomes produced by Generative AI). This work was supported by the Commercialization Promotion Agency for R\&D Outcomes(COMPA) grant funded by the Korea government(Ministry of Science and ICT)(2710096072).

\bibliography{references}

\newpage
\clearpage
\appendix
\begin{table*}[t!]
\centering
\small
\setlength{\tabcolsep}{4pt}
\renewcommand{\arraystretch}{1.18}
\resizebox{\textwidth}{!}{%
\begin{tabular}{>{\raggedright\arraybackslash}p{0.12\textwidth}>{\raggedright\arraybackslash}p{0.18\textwidth}>{\raggedright\arraybackslash}p{0.64\textwidth}}
\toprule
\textbf{Scope} & \textbf{Setting} & \textbf{Implementation details} \\
\midrule
\multirow{2}{*}{\textbf{Shared}}
& Engine / hardware & All generation and PPL computation use vLLM v0.8.5 with bfloat16 precision. Each model pair is served on two NVIDIA A100-80GB GPUs. \\
& Generation budget / sampling & Standard decoding uses \texttt{max\_model\_len}=16384, \texttt{max\_new\_tokens}=16384, temperature=0.6, top-$p$=0.95, top-$k$=-1, and min-$p$=0.0. The standalone Qwen3-1.7B alone uses top-$k$=20 following the Qwen3 recommendation. Within this budget, explicit thinking-stage generation uses an additional 2K-token cap. Self-consistency uses separate sampling settings. \\
\midrule
\multirow{3}{*}{\textbf{\thinkfuse{}}}
& Trigger signal & Every time the primary model generates $T=8$ tokens, we compute top-5 token entropy and aggregate token-level uncertainty into segment-level uncertainty using EWCA with $(\alpha,\beta)=(0.1,0.5)$. \\
& Adaptive threshold & The EMA update rate is $\eta=0.05$, the uncertainty tolerance is $\tau=1.0$, and the initial statistics are $(\mu_0,\sigma_0)=(0.5,0.1)$. The soft-budget coefficient is $\lambda=50$; the trigger uses both recent uncertainty history and the cumulative fusion ratio. \\
& Segment selection & When fusion is triggered, primary and auxiliary candidate segments are compared under the current context using primary-model PPL. Segment boundaries from different tokenizers are aligned by the shared prefix that decodes identically under both tokenizers, and each segment is restored to the reasoning-tag format of the model used for PPL scoring. \\
\midrule
\multirow{4}{*}{\textbf{Baselines}}
& Cool-Fusion & Each participating LLM generates a candidate segment; all generated candidates are scored by the PPL of all participating LLMs, and the candidate with the lowest average PPL is selected. \\
& AdaFuse & The confidence threshold is $\tau_{\Delta}=0.7$, the EMA decay is 0.99, and token confidence is computed from top-5 log probabilities. When the uncertain-token ratio exceeds 0.5, AdaFuse compares $B=2$ candidates. \\
& Self-consistency & We generate $K\in\{3,5\}$ sampled reasoning paths from Qwen3-4B with temperature=0.7 and top-$p$=0.95. Final answers are normalized with the same answer-normalization procedure as the main evaluation and selected by strict majority. If no normalized answer obtains a strict majority, we use the first generated sample. \\
& Post-hoc critique & This appendix-only comparison uses a 16K-token budget. When the primary and auxiliary answers disagree, the auxiliary model critiques the primary output, and the primary model generates a revised answer conditioned on that critique. \\
\bottomrule
\end{tabular}
}
\caption{Implementation hyperparameters. Shared implementation settings and method-specific settings used in the main and appendix experiments.}
\label{tab:impl_hparams}
\end{table*}

\section{Detailed Experimental Setup}
\label{app:imp_details}
\subsection{Hyperparameter settings}
Table~\ref{tab:impl_hparams} summarizes the main implementation settings fixed across the main and appendix experiments. Shared rows report settings used across methods, while method-specific rows report the additional settings required by each baseline or comparison method.

\subsection{Uncertainty Signals and Budget-Penalty Details}
\label{app:uncertainty}

We also consider a confidence-gap variant in ablations:
\begin{equation}
u_{k,i}^{\mathrm{gap}} = 1 - \left(p_{k,i}^{(1)} - p_{k,i}^{(2)}\right),
\end{equation}
where $p_{k,i}^{(1)}$ and $p_{k,i}^{(2)}$ denote the largest and second-largest probabilities among the top-5 candidate tokens at position $i$. This variant assigns higher uncertainty when the two leading candidates have similar probabilities.

The no-budget penalty variant keeps the default entropy-based uncertainty signal, $T{=}8$, and primary-model PPL scoring, but sets only the soft-budget coefficient in the adaptive threshold to $\lambda=0$. Thus, the threshold becomes $\theta_k=\mu_{k-1}+\tau\sigma_{k-1}$, and fusion is triggered only when $U_k^p \ge \theta_k$. As the cumulative fusion ratio $R_{k-1}$ increases, the threshold no longer receives the additional penalty term. This ablation does not fuse every segment; it preserves the EMA-based uncertainty trigger and removes only the soft-budget penalty.

\subsection{Cross-Tokenizer Alignment}
\label{app:alignment}

When the primary and auxiliary models use different tokenizers, we apply shared-prefix text alignment to candidate segment boundaries. We first decode the $\segpreview$-token candidate segment into text, re-encode it with both tokenizers, find the longest prefix whose decoded text is identical under both tokenizers, and use this aligned prefix for segment comparison. In practice, this alignment reduces the effective segment length by 1--2 tokens on average.

\textbf{Reasoning-tag normalization.} Different models use different reasoning tags; for example, Qwen3 uses \texttt{<think>}, whereas EXAONE-Deep uses \texttt{<thought>}. We normalize tags to \texttt{<think>} for fusion and restore them to each model's native reasoning-tag format before PPL computation.

\subsection{Benchmark sampling protocol}
\label{app:sampling}
Due to computational cost constraints, we report all main, analysis, and ablation results on a fixed random subsample of each benchmark rather than the full evaluation split. Evaluating \thinkfuse{} and the baselines across multiple primary-auxiliary model-pair configurations, fusion variants, and ablations on every full benchmark would incur prohibitive compute cost, especially given that the framework involves repeated PPL scoring and parallel auxiliary generation. 
To keep the evaluation tractable while maintaining a consistent comparison across methods, we evaluate each method on a fixed random subsample of 200 examples per benchmark. The subsample is drawn once per benchmark using Python's default  \texttt{random} library generator with seed \texttt{42}, and the same 200 examples are used across all methods, model pairs, and ablations. For MATH-500, GSM8K, and NQ-Open, this corresponds to a strict subsample of the full evaluation split. AIME 2024 (30 problems) and GPQA-Diamond (198 problems) are evaluated in full. All accuracy values reported in the main results, analysis, and ablation tables are computed over the same fixed subsample for each benchmark, ensuring that absolute and relative comparisons between methods are not affected by between-sample variance.

\section{Bootstrap Confidence Intervals}
\label{app:reliability}
Table~\ref{tab:bootstrap_ci} reports 95\% bootstrap confidence intervals for the accuracy delta of \thinkfuse{} over the standalone Qwen3-4B on each benchmark. The deltas are computed from re-executed runs, so point estimates may differ slightly from Table~\ref{tab:main_results}. All twenty pair-benchmark deltas are positive, eleven intervals exclude zero, and no interval indicates significant degradation. The wide AIME24 intervals reflect its small sample size of 30 problems.

\begin{table*}[t!]
\centering
\small
\setlength{\tabcolsep}{3pt}
\renewcommand{\arraystretch}{1.1}
\begin{tabular}{lccccc}
\toprule
\textbf{Auxiliary} & \textbf{MATH-500} & \textbf{GSM8K} & \textbf{AIME24} & \textbf{GPQA} & \textbf{NQ-Open} \\
\midrule
Ministral & +0.5 [$-$1.5, +2.5] & +3.1 [+1.0, +5.6]* & +20.0 [+6.7, +36.7]* & +10.6 [+1.0, +20.2]* & +10.0 [+5.5, +15.0]* \\
EXAONE-Deep & +2.0 [+0.5, +4.0]* & +3.1 [+1.0, +5.8]* & +20.0 [+6.7, +33.3]* & +1.0 [$-$8.1, +10.1] & +9.0 [+4.5, +14.0]* \\
DeepSeek-R1 & +1.5 [$-$0.5, +4.0] & +3.0 [+0.0, +6.5] & +3.3 [$-$10.0, +16.7] & +8.7 [$-$0.5, +17.9] & +8.0 [+3.5, +13.1]* \\
Qwen3-1.7B & +2.0 [+0.5, +4.0]* & +2.0 [$-$0.5, +5.0] & +16.7 [$-$3.3, +36.7] & +1.0 [$-$8.6, +10.1] & +7.5 [+3.5, +12.0]* \\
\bottomrule
\end{tabular}
\caption{95\% bootstrap confidence intervals for the accuracy delta of \thinkfuse{} over the standalone Qwen3-4B with each auxiliary. * marks intervals excluding zero. Deltas are from re-executed runs and may differ slightly from Table~\ref{tab:main_results}.}
\label{tab:bootstrap_ci}
\end{table*}

\begin{table*}[t!]
\centering
\small
\setlength{\tabcolsep}{4pt}
\resizebox{0.9\textwidth}{!}{%
\begin{tabular}{lccccc}
\toprule
\textbf{Method} & \textbf{MATH-500} & \textbf{GSM8K} & \textbf{AIME24} & \textbf{GPQA} & \textbf{NQ-Open} \\
\midrule
\multicolumn{6}{l}{\emph{Standalone models}} \\
Qwen3-4B  & 89.0 & 92.0 & 53.3 & 54.0 & 29.5 \\
EXAONE-Deep & 83.5 & 90.5 & 43.3 & 57.1 & 17.5 \\
DeepSeek-R1  & 13.0 & 30.5 & 0.0\truncmark  & 21.2 & 14.5 \\
\midrule
\multicolumn{6}{l}{\emph{\thinkfuse{} cross-family pairs}} \\
Qwen3-4B $\times$ EXAONE-Deep  & \textbf{91.0}\deltagain{+2.0} & \textbf{99.0}\deltagain{+7.0} & \textbf{73.3}\deltagain{+20.0} & 55.1\deltagain{+1.1} & \textbf{38.5}\deltagain{+9.0} \\
Qwen3-4B $\times$ DeepSeek-R1 & 90.5\deltagain{+1.5} & 95.0\deltagain{+3.0} & 56.7\deltagain{+3.4} & \textbf{63.3}\deltagain{+9.3} & 37.7\deltagain{+8.2} \\
\bottomrule
\end{tabular}
}
\caption{Additional \thinkfuse{} model pairs. Accuracy (\%) for the cross-family pairs with EXAONE-Deep and DeepSeek-R1 as auxiliary models under the same 16K-token budget and evaluation protocol as Table~\ref{tab:main_results}. Values in parentheses indicate the signed accuracy difference relative to the standalone Qwen3-4B. The $\dagger$ symbol marks AIME24 results affected by length-limit or final-answer truncation.}
\label{tab:full_results}
\end{table*}

\section{Additional Model-Pair Results of \thinkfuse{}}
\label{app:full_results}
Table~\ref{tab:full_results} reports the \thinkfuse{} configurations that are not included in Table~\ref{tab:main_results}, namely the cross-family pairs with EXAONE-Deep (EXAONE-Deep-2.4B) and DeepSeek-R1 (DeepSeek-R1-Distill-Qwen-1.5B) as auxiliary models. Both pairs follow the same evaluation protocol as the main experiments and improve over the standalone Qwen3-4B on every benchmark, matching the pattern of the pairs reported in the main text.

\section{Additional Sampling Diagnostic}
\label{app:sampling_diagnostic}
Table~\ref{tab:app_qwen_pass5} reports the oracle pass@5 upper bound obtained by independently sampling Qwen3-4B five times. Unlike self-consistency, this diagnostic counts an instance as correct if any of the five sampled responses contains the correct answer. We therefore use it only as an auxiliary diagnostic of the potential upper bound from repeated sampling, not as a main baseline.

\begin{table}[t]
\centering
\small
\setlength{\tabcolsep}{6pt}
\resizebox{0.8\columnwidth}{!}{%
\begin{tabular}{lc}
\toprule
\textbf{Benchmark} & \textbf{Qwen3-4B pass@5} \\
\midrule
MATH-500 & 90.5\deltagain{+1.5} \\
GSM8K & 95.5\deltagain{+3.5} \\
AIME24 & 70.0\deltagain{+16.7} \\
GPQA & 71.2\deltagain{+17.2} \\
NQ-Open & 48.5\deltagain{+19.0} \\
\bottomrule
\end{tabular}
}
\caption{Qwen3-4B pass@5 oracle sampling upper bound.
\texttt{pass@5} counts an instance as correct if at least one of five independently sampled Qwen3-4B responses contains the correct answer. Values are accuracy (\%); colored values in parentheses denote the difference from Qwen3-4B standalone accuracy. This table reports only the sampling upper-bound diagnostic, which is not included in the main results table.}
\label{tab:app_qwen_pass5}
\end{table}

\section{Fusion position distribution by correctness}
\begin{figure}[t]
\centering
\includegraphics[width=0.95\columnwidth]{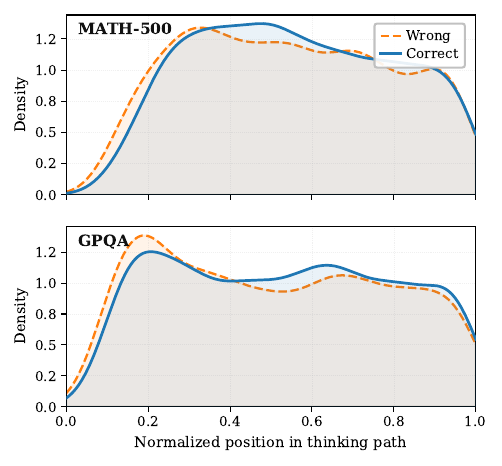}
\caption{\thinkfuse{} fusion-position distribution. Normalized positions of thinking-stage segments where fusion is triggered, separated by final-answer correctness. The primary model is Qwen3-4B and the auxiliary model is Ministral-3B-Reasoning.}
\label{fig:fusion_position}
\end{figure}

Figure~\ref{fig:fusion_position} shows the normalized positions of thinking-stage segments where fusion is triggered in \thinkfuse{}, separating examples by final-answer correctness. The primary model is Qwen3-4B and the auxiliary model is Ministral-3B-Reasoning. Position 0 denotes the point immediately after \texttt{<think>}, and position 1 denotes the point immediately before \texttt{</think>}. In MATH-500, fusion for incorrectly answered examples is more dispersed across early and middle parts of the thinking stage, whereas GPQA shows a smaller difference by final-answer correctness.

\section{Comparison with Post-Hoc Critique}
\label{app:posthoc_critique}

\thinkfuse{} selects between primary and auxiliary candidate segments at uncertain segment boundaries during the thinking stage using primary-model PPL. For comparison, we also evaluate a post-hoc critique baseline. In this baseline, the primary and auxiliary models first generate complete reasoning trajectories and answers independently. If the two answers agree, we keep the primary output; otherwise, the auxiliary model critiques the primary output, and the primary model produces a revised answer conditioned on the critique. Table~\ref{tab:posthoc_critique} compares the two approaches with the same Qwen3-4B $\times$ EXAONE-Deep-2.4B model pair and the same evaluation protocol as the main experiments.

\begin{table}[t]
\centering
\small
\setlength{\tabcolsep}{5pt}
\resizebox{\columnwidth}{!}{%
\begin{tabular}{lcc}
\toprule
\textbf{Benchmark} & \textbf{\thinkfuse{}} & \textbf{Post-hoc critique} \\
\midrule
MATH-500 & \textbf{91.0}\deltagain{+2.0} & 86.3\deltaloss{$-$2.7} \\
GSM8K  & \textbf{99.0}\deltagain{+7.0} & 92.0\deltaneutral{+0.0} \\
AIME24 & \textbf{73.3}\deltagain{+20.0} & 30.0\deltaloss{$-$23.3} \\
GPQA & \textbf{55.1}\deltagain{+1.1} & 27.8\deltaloss{$-$26.2} \\
NQ-Open  & 38.5\deltagain{+9.0} & \textbf{40.0}\deltagain{+10.5} \\
\bottomrule
\end{tabular}
}
\caption{\thinkfuse{} vs. post-hoc critique. Accuracy (\%) with the same Qwen3-4B $\times$ EXAONE-Deep-2.4B model pair and the same evaluation protocol as the main experiments. Colored values in parentheses denote the difference from Qwen3-4B standalone accuracy.}
\label{tab:posthoc_critique}
\end{table}

\paragraph{Limitations of Post-Hoc Critique.}
Post-hoc critique intervenes only after the primary model has completed its reasoning trajectory. It therefore re-evaluates and revises a completed solution rather than replacing an incorrect intermediate branch when it first appears. In contrast, \thinkfuse{} intervenes locally at segment boundaries during the thinking stage, preserving reusable reasoning context while replacing only the uncertain segment.

\section{Supplementary Comparisons with Uncertainty-Guided Test-Time Scaling Methods}
\label{app:related_comparison}
MUR~\cite{yan-etal-2026-mur} and SpecReason~\cite{specreason} guide test-time scaling with uncertainty, as \thinkfuse{} does, but pursue different objectives. MUR scales computation within a single model, and SpecReason accelerates a large model through speculative step drafting. We examine how these uncertainty-based intervention policies behave compared to \thinkfuse{} to further isolate the effect of trajectory-calibrated fusion triggering.

\subsection{Comparison with MUR}
\label{app:mur}
MUR tracks momentum-smoothed step-level uncertainty and allocates additional computation to uncertain steps within a single model. We re-implement MUR from the official repository with Qwen3-4B under the 16K-token budget and evaluate two variants, the default $\phi$-decoding and a MUR-style variant that employs Ministral-3B-R as a critic. As shown in Table~\ref{tab:mur_comparison}, both variants fall below the standalone Qwen3-4B, whereas \thinkfuse{} improves it on both benchmarks. We attribute this gap in part to MUR's prompted step segmentation, which inserts explicit step markers into the trace and alters the natural reasoning flow over which uncertainty is measured. \thinkfuse{} instead monitors fixed-length segments without modifying the generation format.

\begin{table}[t!]
\centering
\small
\setlength{\tabcolsep}{5pt}
\renewcommand{\arraystretch}{1.08}
\begin{tabular}{@{}lcc@{}}
\toprule
\textbf{Method} & \textbf{AIME24} & \textbf{GPQA} \\
\midrule
Qwen3-4B standalone & 53.3 & 54.0 \\
MUR ($\phi$-decoding) & 16.7\deltaloss{$-$36.6} & 37.9\deltaloss{$-$16.1} \\
MUR (Ministral critic) & 53.3\deltaneutral{+0.0} & 48.0\deltaloss{$-$6.0} \\
\midrule
\thinkfuse{} & \textbf{73.3}\deltagain{+20.0} & \textbf{64.6}\deltagain{+10.6} \\
\bottomrule
\end{tabular}
\caption{Comparison with MUR. Accuracy (\%) with Qwen3-4B as the reasoning model under the 16K-token budget and the same evaluation protocol as the main experiments. \thinkfuse{} pairs Qwen3-4B with Ministral-3B-R. Values in parentheses denote the difference from Qwen3-4B standalone accuracy.}
\label{tab:mur_comparison}
\vspace{1.0em}

\centering
\small
\setlength{\tabcolsep}{5pt}
\renewcommand{\arraystretch}{1.08}
\begin{tabular}{@{}lcc@{}}
\toprule
\textbf{Method} & \textbf{AIME24} & \textbf{GPQA} \\
\midrule
Qwen3-4B standalone & 53.3 & 54.0 \\
DeepSeek-R1 standalone & 0.0\truncmark & 21.2 \\
SpecReason & 0.0\deltaloss{$-$53.3} & 44.4\deltaloss{$-$9.6} \\
\midrule
\thinkfuse{} & \textbf{56.7}\deltagain{+3.4} & \textbf{63.3}\deltagain{+9.3} \\
\bottomrule
\end{tabular}
\caption{Comparison with SpecReason. Accuracy (\%) under the 16K-token budget and the same evaluation protocol as the main experiments. SpecReason uses DeepSeek-R1-Distill-Qwen-1.5B as the generator and Qwen3-4B as the verifier, and \thinkfuse{} uses the same two models as primary and auxiliary. Values in parentheses denote the difference from Qwen3-4B standalone accuracy. The $\dagger$ symbol marks AIME24 results affected by length-limit or final-answer truncation.}
\label{tab:specreason_comparison}
\end{table}

\subsection{Comparison with SpecReason}
\label{app:specreason}
SpecReason lets a small model draft each reasoning step and a larger base model score every draft through prompting-based critique, accepting the draft or regenerating the step itself. We run the public implementation with DeepSeek-R1-Distill-Qwen-1.5B as the generator and Qwen3-4B as the verifier, keeping the acceptance threshold of 7, greedy verifier scoring, and the 16K-token budget. As shown in Table~\ref{tab:specreason_comparison}, SpecReason improves over its standalone generator but remains below both the standalone Qwen3-4B and \thinkfuse{}. The verifier fallback operates as intended, accepting only 15.6\% of the proposed steps on GPQA, and most AIME24 failures end in repetitive reasoning without a valid answer rather than in budget truncation. These results reflect the different goals of the two methods. SpecReason approximates the quality of its base model at lower cost by verifying every step, whereas \thinkfuse{} calibrates intervention against the trajectory history and targets accuracy beyond the stronger model.

\end{document}